\documentclass[letterpaper, 10 pt, conference]{ieeeconf}  % Comment this line out
\IEEEoverridecommandlockouts                              % This command is only
\usepackage{amsmath,amsfonts}
\usepackage{algorithmic}
\usepackage{algorithm}
\usepackage{array}

\usepackage{textcomp}
\usepackage{stfloats}
\usepackage{url}
\usepackage{verbatim}
\usepackage{graphicx}
\usepackage{cite}

\usepackage{amssymb}
\usepackage{siunitx}

\usepackage[hidelinks]{hyperref}
\usepackage{booktabs}
\usepackage{multirow}

\usepackage{balance}

\usepackage{caption}
\usepackage{subcaption}
\usepackage{soul}
\newcommand{\ch}[1]{\textcolor{black}{#1}}

\title{Propeller Damage Detection: Adapting Models to Diverse UAV Sizes}

\author{Gabriel Torre*, Claudio Pose and Juan Giribet
\thanks{\copyright~© 20XX IEEE.  Personal use of this material is permitted.  Permission from IEEE must be obtained for all other uses, in any current or future media, including reprinting/republishing this material for advertising or promotional purposes, creating new collective works, for resale or redistribution to servers or lists, or reuse of any copyrighted component of this work in other works.
Accepted version. Published in: 2025 International Conference on Unmanned
Aircraft Systems (ICUAS), pp.~534--541, 2025.
DOI: \href{https://doi.org/10.1109/ICUAS65942.2025.11007793}{10.1109/ICUAS65942.2025.11007793}}
\thanks{Gabriel Torre is with Universidad de Buenos Aires, Instituto de Ingeniería Biomédica, Facultad de Ingeniería and Universidad de San Andrés, Laboratorio de Inteligencia Artificial y Robótica  {\tt\small torreg@udesa.edu.ar}}
\thanks{Claudio Pose is with Laboratorio de Automática y Robótica, Facultad de Ingeniería, Universidad de Buenos Aires and CONICET - Universidad de San Andrés {\tt\small cldpose@fi.uba.ar}}
\thanks{Juan Giribet is with  Laboratorio de Inteligencia Artificial y Robótica, Universidad de San Andrés and CONICET {\tt\small jgiribet@conicet.gov.ar}}
}

\begin{document}

\maketitle
\thispagestyle{empty}
\pagestyle{empty}

\begin{abstract}

\ch{This manuscript introduces a transfer learning method for adapting propeller fault detection neural networks to different unmanned aerial vehicles (UAVs). After training a simple model for detecting if any propeller in a specific vehicle has a failure (in this case, a chipped tip), a domain adaptation based in the vehicles' physics is performed in order to use the same model to detect failures in vehicles with different structures, weights, or motor-propeller sets. A key feature is that the detection model uses only inertial sensors that are standard in commercial UAVs, making it broadly applicable without the need for additional hardware.}
\end{abstract}

\section*{Supplementary Material}
The code and data are available at \url{https://github.com/udesa-ai/prop-damage-adapt} and the flight video at \url{https://www.youtube.com/watch?v=MXNc1VfQ0Y8}

%% main text
\section{Introduction}\label{intro}

Unmanned Aerial Vehicles (UAVs) have become ubiquitous due to their versatility, low cost, and ease of use in various applications such as agriculture, delivery, and aerial photography. As UAVs increasingly interact with the public in both open and closed spaces, ensuring their safe operation becomes paramount.

This paper addresses one of the most common and critical UAV failures: propeller damage in the actuator subsystem. In addition to electric motors, propellers are highly exposed to environmental conditions, which can significantly degrade their performance, particularly when operating in sandy, wet, or harsh environments. The actuator subsystem (excluding the electronic speed controller of the motors) requires the most maintenance, as propellers need to be replaced after impacts and motors often require bearing changes after a given number of flight hours.

Damaged propellers introduce vibrations into the UAV structure, which are detected by inertial sensors. These vibrations can make those measurements unusable and potentially lead to a loss of stability. Additionally, damaged propellers reduce actuator efficiency, lowering thrust, and increasing power consumption and operating temperature.

\begin{figure}[t]
  \centering
  \includegraphics[width=\linewidth]{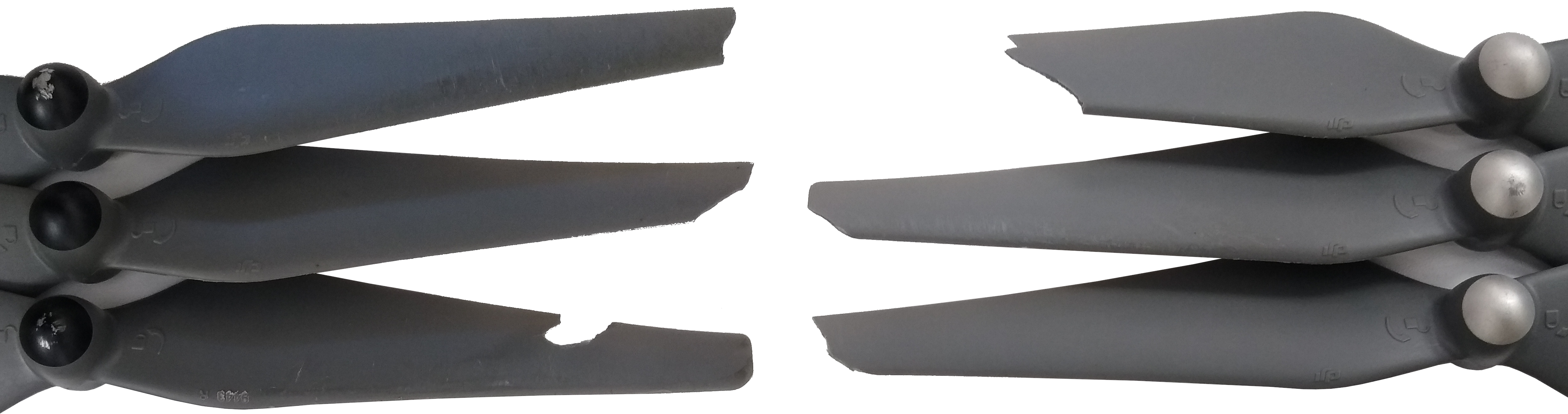}
  \caption{Propellers damaged during normal operation.}
  \label{fig:helices}
\end{figure}

Over time, propellers can sustain progressive damage, with blade tips particularly prone to chipping, as shown in Fig.~\ref{fig:helices}. Notably, even the Mars helicopter Ingenuity, developed by NASA's Jet Propulsion Laboratory, experienced propeller tip damage, which ultimately led to the mission end \cite{nasaAfterThree}. Detecting such damage, especially when not visually apparent, is crucial for safe UAV operation. 
Although propeller tip damage does not necessarily lead to immediate catastrophic failure, it degrades performance. This forces motors to operate closer to their limits, thereby shrinking the flight envelope and making the vehicle progressively less tolerant to external perturbations, such as wind, which compromises safety.

To ensure vehicle safety and reliability, it is vital to identify these anomalies promptly to enable proactive maintenance or take countermeasures during flight if critical damage occurs. While some kind of direct measurement would be ideal, in practice it is difficult to perform due to the rotating propeller, so several direct and indirect methods have been proposed to estimate its state \cite{You2024,Palanisamy2022,Canle2019,Hylmar2024}. However, achieving this is challenging, as damage effects on the structure can vary depending on factors such as propeller speed and environmental conditions. 

Still, the shape, size, and nature of the damage can take numerous forms, making it impossible to accurately model their effects using traditional mathematical modeling. Therefore, data-driven methods offer a more effective solution \cite{OConnell2024}.
Recent success in spectral features, which capture key information in the frequency domain, has demonstrated the potential of data-driven approaches to detect propeller damage in UAVs \cite{Wang2024, pose2023, Baldini2023, pose2025}. This work leverages these methods to detect and characterize propeller damage.

Despite their advantages, data-driven solutions have limitations. They depend heavily on the data with which they are trained, and to generalize to a broader dataset, the training data must be carefully selected. Even then, additional considerations may be needed when applying trained models to new datasets. In this case, the structure of the vehicle, the materials it is made from, the characteristics of the motors and propellers, and the types of sensors onboard can significantly affect the data collected to detect damaged propellers.

What makes the problem challenging is the subtlety and variety of propeller damage manifestations, which often do not lead to immediate, observable failures but gradually impair the UAV's performance. If a data-based approach is to be used, it would require collecting data both for a healthy vehicle and for a wide array of representative types of propeller damage. 
% This has been an active topic over the past few years. 
However, applying this solution to a different vehicle may severely degrade its performance, requiring another full dataset for the new vehicle considering all the same types of damaged propellers.

Previous methods for detecting UAV propeller failure have been explored \cite{Ghalamchi2018,pose2023,Etchemaite2023}, but often struggle to generalize across different vehicle domains or load profiles. This limitation underscores the challenge of adapting these methods to varied UAV configurations without compromising their effectiveness.

A range of indirect measurement techniques have been proposed, including analysis of inertial and control data \cite{Fu2019}, PWM signals \cite{Baldini2023}, propeller RPM \cite{Saied2017}, current sensors \cite{Pourpanah2018}, and acoustic signatures \cite{Iannace2019,Bondyra2022,Rangel2018}. However, the reliance on specific sensors that are not universally available across all UAV models introduces a significant barrier to the widespread application of these methods.

Furthermore, most existing approaches \cite{Bondyra2018,Ghalamchi2020} require that the detection system be tailored to the exact vehicle configuration for which damage detection is intended. This specificity greatly restricts the practical utility of these solutions, as it prevents their application in diverse operational contexts. Then, even if the data-based approach proves successful, the solution is mostly limited to the vehicle used to collect the data, or to slightly different configurations. The power spectra of the inertial sensors is primarily defined by the vibrations introduced by the rotating propellers and mechanical properties of the structure, which can also be affected by existing physical dampeners or software filters. For example, a variation in weight, which requires more thrust, and thus higher speed of rotation from the propellers, may substantially change the power distribution of the vibrations and render the data-based solution unusable.

Notably, transfer learning is showing promise in UAV propeller damage detection. For example, \cite{liu2020} shows how an acoustic-based damage detection model trained on one UAV type can be adapted to another by fine-tuning with a limited data set. A similar approach using sensor data is applied in \cite{Liu2023} to detect actuator failures in fixed-wing UAVs.

\ch{To overcome the limitation of these solutions to be only applicable to a specific vehicle, our proposal is to define a way to apply a successfully trained damage detection network obtained for a specific UAV, to a different vehicle, considering that the vehicles are at least of the same UAV type (i.e. both are quadrotors, or the same quadrotor with different characteristics, etc.). }

To this end, this manuscripts presents a method that adapts the data acquired by the second vehicle to be used in the trained solution of the first vehicle, based only in the physical relationships between both.

%\medskip
The contributions of this work are the following:
\begin{itemize}
\item Introduce a method based on first principles for adapting damage detection models to different UAVs without re-training.

\item Show how variations in physical attributes such as weight and UAV configuration (e.g., quadrotors versus hexarotors) influence the power spectra captured by a UAV's inertial sensors, revealing different sensitivities to propeller damage.

\item Enhance the computational efficiency of an existing propeller damage detection algorithm.

\item \ch{Propose improvements over a previous version, enhancing computational efficiency of an existing propeller damage detection algorithm, in views of a future real-time implementation.}

\end{itemize}

While in the following sections the proposed method will be evaluated using one particular type of propeller damage, previous work from the authors such as \cite{pose2025} shows that several types of damage can be detected, isolated, located and identified using similar architectures as those proposed here. However, this work focuses on the ability to reuse damage detection model trained in one vehicle on a different one, and for the sake of brevity a single type of damage is used. Future work will extend the results for a wider range of damage types.

\section{Method}

\ch{The proposed workflow begins by training a simple propeller damage detection system for a specific vehicle type, which in this case will be a standard quadrotor. Here, a single neural network will be used for detecting if a propeller is affected by a chipped tip, for a particular type of damage. Then, the exact same network is used to detect the same kind of failure in different vehicles, by using a domain adaptation method. This approach is particularly advantageous for adapting the damage detection system to various load configurations, a feature crucial for some UAVs such as those used for transport and delivery and which may carry different weights on each flight, or for agricultural spraying UAVs where the weight changes continuously during a flight as tanks deplete. However, it is essential to note that the source and target UAVs must share a degree of similarity for the transfer process to be successful. As it will be shown later, transferring the detection system  from a quadrotor to a hexarotor resulted in a sub-optimal solution, highlighting a limitation of this method.}

\ch{The proposed neural network architecture processes spectral features from inertial sensor data, and is trained using flight data recorded with the quadrotor vehicle structure. The network output is used as a classifier, to detect if the vehicle is healthy or presenting a damaged propeller. This network, as is, is then applied to detect damage in different UAV structures which are mainly of the same type (quadrotors) but with differences in frame size, motors, propellers, weight and others. It will also be tested with a different structure (hexarotor) but with sub-optimal results, showing the limitations of the method.}

\ch{The proposed solution is possible through domain adaptation, by adjusting the features extracted from the data recorded on each new vehicle, to an equivalent of the features extracted from the original vehicle. The adjustment is done by accounting for differences in weight and propeller characteristics between the original and target vehicles, and by using a brief sample of nominal flight data to identify a characteristic spectral shift.}

\ch{While the damage detection network is simple and could be implemented by other means, the goal in this work is to show that such a solution can be adapted for several vehicles, saving time in re-training, and avoiding performing flights with damaged propellers in new vehicles. In future work, it is intended to do a similar process with more complex networks for estimating damage, such as the ones presented in \cite{pose2023}.}

\section{Damage Detection}\label{daclass}

\begin{figure}[t]
  \centering
  \includegraphics[width=\linewidth]{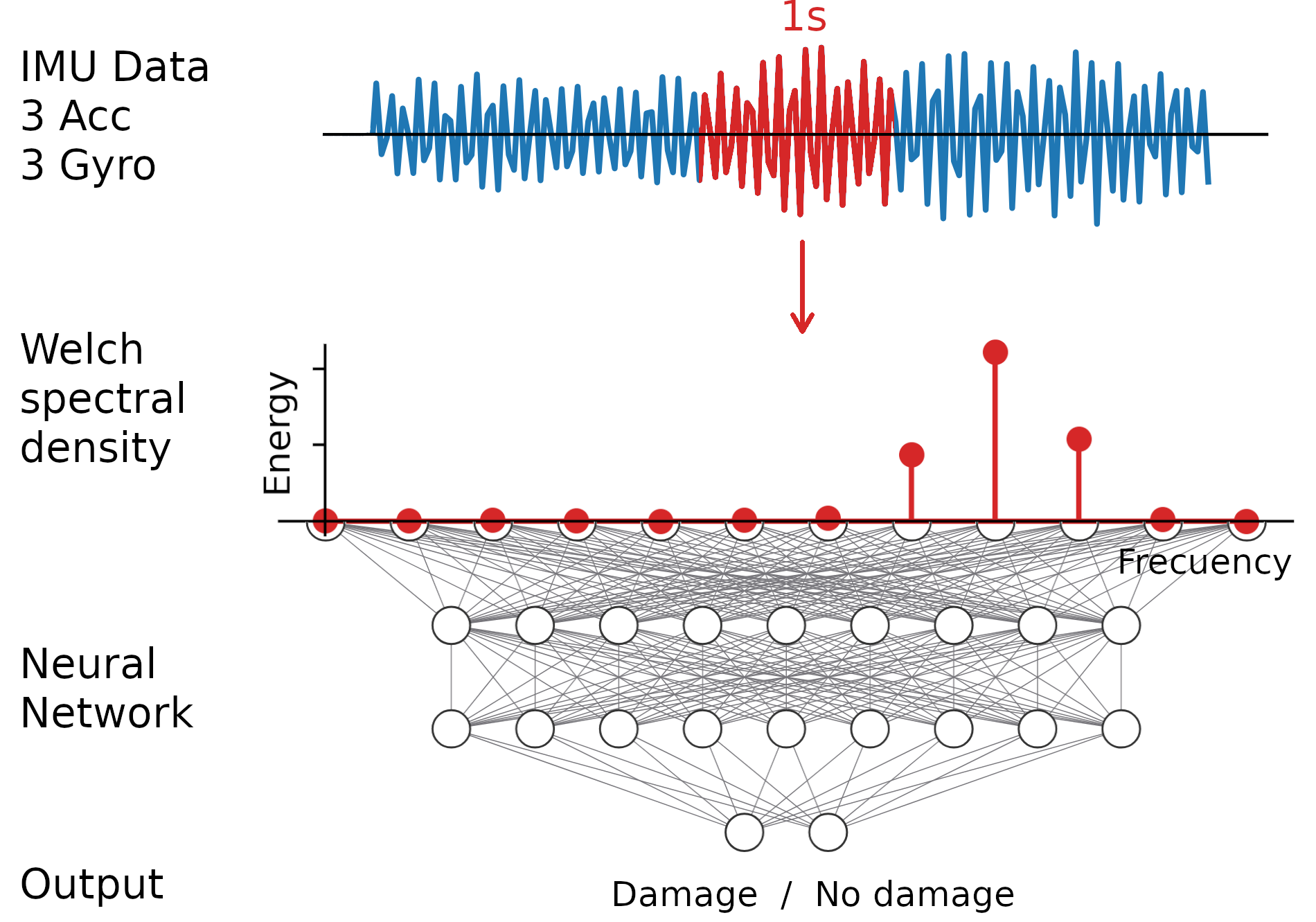}
  \caption{Architecture of the propeller damage detection system. Inertial data is processed into spectral features using the Welch method, which is then used as input into a neural network to predict the existence of damage.}
  \label{fig:class}
\end{figure}

\ch{The damage detection approach proposed in this work is a simplified version of the model used in \cite{pose2023}, with some key differences. In this manuscript, a less complex approach is implemented, where a single neural network functions as a binary classifier capable of detecting whether the vehicle is damaged or in a healthy state. The diagram in Fig.~\ref{fig:class} summarizes the architecture used and which will be described in detail in the following subsections.}

\subsection{Sensors}

In order to generate data that is meaningful for propeller damage detection, it will be obtained exclusively from an Inertial Measurement Unit (IMU), which consists of triaxial accelerometers and gyroscopes. As propeller damage produces vibrations consistent with the propeller's rotational speed, this sensor provides valuable information about the state of the vehicle. Furthermore, these sensors are essential for every UAV, as they provide the minimum data required to estimate the vehicle's attitude, which is critical for its stability. Consequently, such a sensor is always present in any vehicle of this kind, ensuring its availability for damage detection.

The raw data comprises measurements from three accelerometers and three gyroscopes, one for each axis, sampled at a frequency of \SI{222}{\hertz} during flights with either healthy or damaged propellers. The sampling frequency depends on the data acquisition rate of the flight controller used, \ch{in this case a Pixhawk running PX4 firmware,} which gathers and records data at \SI{222}{\hertz} for identification purposes.

\subsection{Data Curation}

\ch{The recorded flight data is curated to remove portions that are not representative of the vehicle’s state, specifically the take-off and landing phases, which introduce various perturbations and ground effects that are not present during the rest of the flight.}

\ch{Afterwards, if the data is used for training a network, it is divided into non-overlapping 10-second sections, which are randomly assigned to either the training or validation sets in a 90\%-10\% proportion. Then, each of these sections is further subdivided into one-second windows with 50\% overlap, from which the input features for the neural network will be generated. The reason for the two-stage partitioning is to avoid having two consecutive one-second windows in the training and validation sets, thus lowering the correlation between sets. For the same reason, a completely separate flight is used for the test set, curated with the same procedure.}

\ch{In the case the data is used for domain adaptation, it is directly partitioned into one-second windows with 50\% overlap.}

\ch{The flight trajectories feature a wide range of maneuvers in each degree of freedom (tilts in pitch and roll, turns in yaw, ascent and descent, straight flight, and all the possible combinations). In the case of flights with a damaged propeller, 
it is always located in the same motor, and subsequently data augmentation is used to simulate damage in each of the other propellers by performing an adequate rotation of the sensor data. This assumes a degree of symmetry in the vehicle's structure and a center of mass located near its geometric center, which is typically true for standard drones and was validated in \cite{pose2023}.}

\subsection{Spectral Representation}
To detect propeller damage, we employ spectral features, which are more effective than time-series data for identifying vibrations caused by rotating propellers, whether healthy or damaged. Our previous work \cite{pose2023} used a filter bank with filters equispaced in frequency and computed the spectral density over each filtered frequency band using FFT. \ch{Here, we instead implement the Welch method \cite{welch}, which is an efficient technique for estimating spectral density that avoids the need for full FFTs, with the same result and much less demanding computationally.}

Each one-second window from the six inertial sensors is processed using the Welch method to estimate the power spectrum in \SI{5}{\hertz} segments, which was found to be optimal for this configuration in \cite{pose2023}. For each sensor, the power density is calculated over 22 segments (sampled at \SI{222}{\hertz}, Nyquist frequency of \SI{111}{\hertz}), yielding a total of $22\times6=132$ features to serve as input for the neural network.

\subsection{Neural Network}
The core of our approach is a neural network consisting of two hidden layers, each with 256 neurons and ReLU activation functions. The model is optimized using the Adam algorithm with a learning rate of 3e-4, early stopping and a 50\% dropout rate is applied for regularization. Several experiments showed that increasing the number of neurons beyond this configuration does not significantly improve accuracy with our dataset~\cite{Data}. However, more expressive models could be beneficial for richer datasets. The network has two outputs: one for the healthy state and one for the damaged state, with a softmax function applied over them. Multiple neurons and a softmax are used, opposed to a single neuron with a sigmoid like activation, to make the method easily extendable to other problem formulations, like the three class problem (healthy, light damage, heavy damaged) formulation used in \cite{Baldini2023}.

\section{Domain Adaptation}\label{domadapt}
To enable the use of models or data from one UAV for damage detection in another, we propose a domain adaptation method based on first principles. A key characteristic of UAV power spectra is a peak corresponding to the propeller rotation speed. For example, in a standard quadrotor, under stable flight conditions, this peak is approximately centered at \(f = F(W/4)\), where \(W\) is the vehicle's weight and \(F(\cdot)\) represents the empirical relationship between frequency and thrust for a specific motor-propeller set. This formalizes the observation that, when all propellers are healthy, each motor-propeller set produces thrust equal to a quarter of the vehicle's weight, with all rotating at the same speed (frequency). Thus, shifts in the frequency spectrum are primarily influenced by the propeller design and the UAV’s weight. We introduce a scaling factor for domain alignment in the frequency domain, defined as in Equation \eqref{eq}.  

\begin{equation}\label{eq}
s_{2 \rightarrow 1} = \frac{F_2(W_2/4)}{F_1(W_1/4)}
\end{equation}

\begin{figure}[t]
  \centering
  \includegraphics[width=\linewidth]{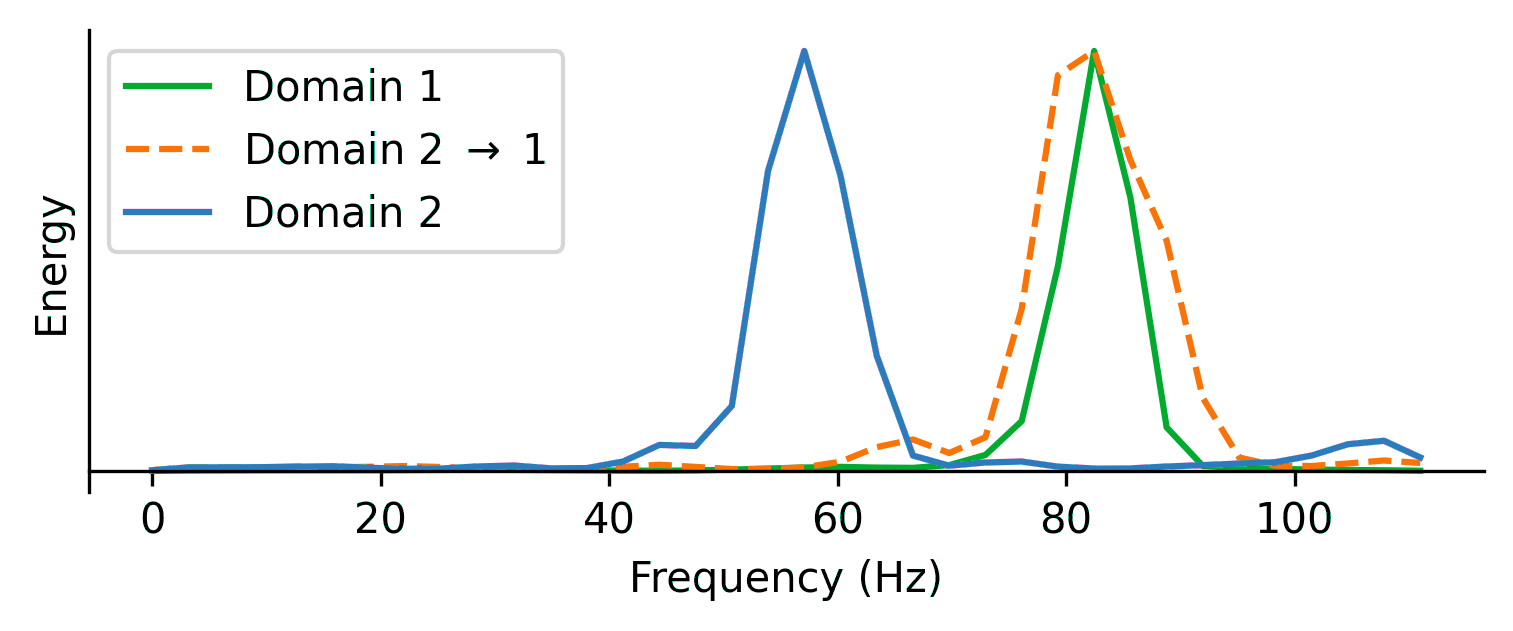}
  \caption{Normalized spectral energy density for the accelerometer $x$ axis for Domains 1 and 2, showing the effect of applying domain alignment.}
  \label{fig:dist}
\end{figure}

Figure~\ref{fig:dist} shows the power density for two different quadrotor UAVs \ch{(Domain 1 and Domain 2), obtained by applying the Welch method to the temporal data recorded during a full flight, where all the propellers were healthy. As it can be noted, each vehicle presents a distinctive peak at \SI{83}{\hertz} and \SI{57}{\hertz} respectively, corresponding to the nominal frequency of the motors in a healthy state.}

\ch{In orange, the impact of domain adaptation is shown, as the peak of Domain 2 is converted and aligned to that of Domain 1}. 
%Building on this understanding of power spectra shifts, we propose a comprehensive domain adaptation workflow. 
This approach enables training a model on one vehicle, where abundant data is available, and deploying it on another vehicle, for which either only propeller test bench data or a brief sample of nominal flight is available, even if no previous propeller damage data exists. \ch{A minimum information about the new vehicle is needed, for example, where the nominal power peak is located, in order to correctly perform the domain adaptation. This can also be done based on bench test data of the motor-propeller sets and knowledge of the weight of the vehicle.}

\begin{figure}[t]
  \centering
  \includegraphics[width=\linewidth]{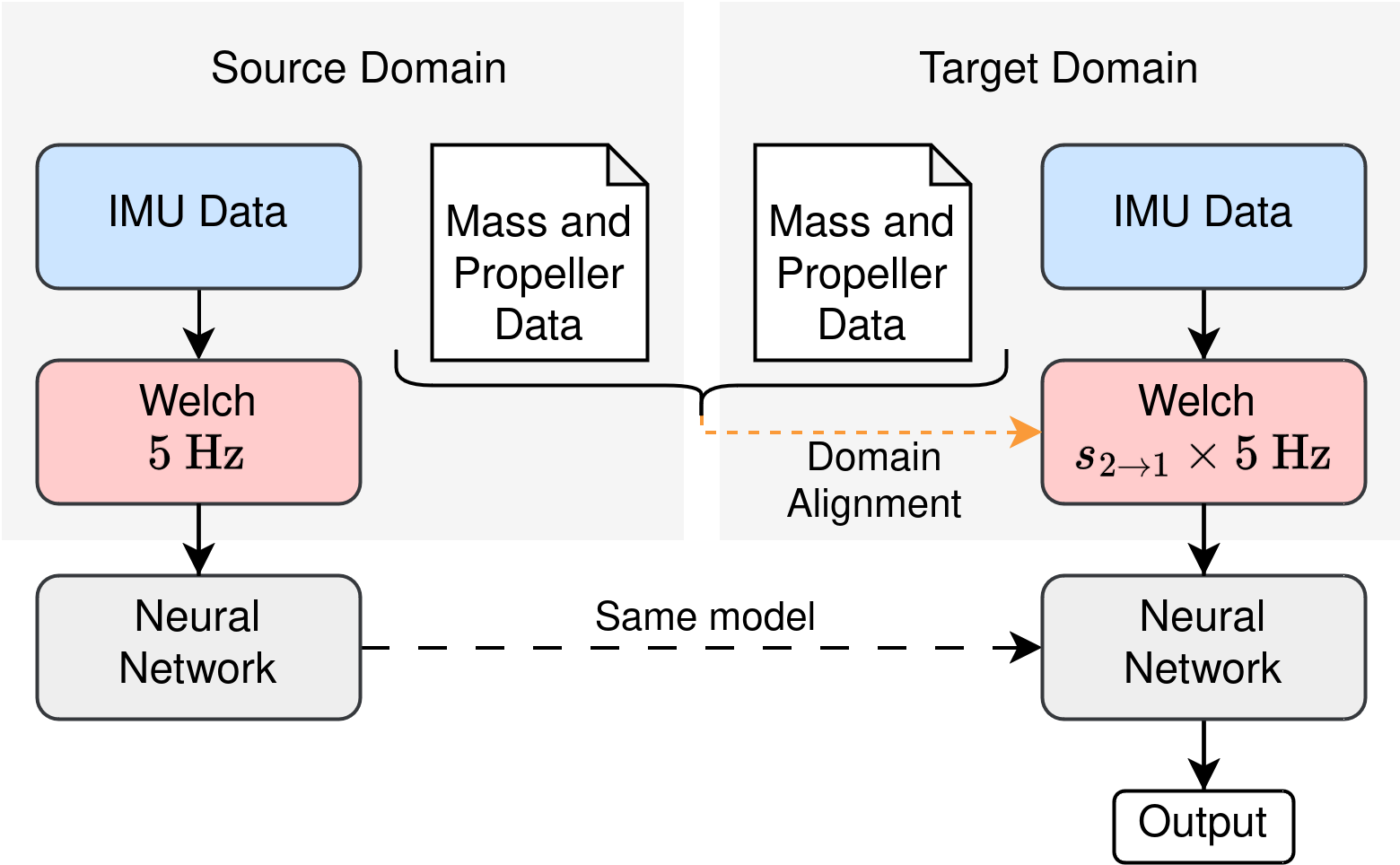}
  \caption{Domain adaptation workflow illustrating how a neural network model, trained on the source domain with simulated propeller damage, can be adapted for the target domain using a calculated domain alignment constant.}
  \label{fig:transport}
\end{figure}

Fig.~\ref{fig:transport} illustrates the application of domain adaptation. The neural network is initially trained on the source domain using the architecture described in Section~\ref{daclass}. To align the target domain with the source, the Welch window sizes are scaled by the factor \(s_{2 \rightarrow 1}\). Specifically, windows of size \(s_{2 \rightarrow 1} \times\)\SI{5}{\hertz} are used and fed into the network as though they were the original \SI{5}{\hertz} windows.

If the scaling factor contracts the domain, any missing values are filled by repeating the last value. In our experiments, using this method or filling the gap with zeros did not show a significant difference in performance. This method assumes the data is sampled at a Nyquist frequency that captures at least one frequency peak generated by the propellers.

This approach is advantageous because it does not rely on the actual power distributions, making it robust against other resonance effects and spectral peaks that might arise.

Alternatively, if no propeller test bench data is available but a brief sample of nominal flight data is provided and the peaks are clearly identifiable, the scale factor \(s_{2 \rightarrow 1}\) can be selected empirically by matching the peak values of both spectra.

\section{Experimental Setup}\label{exp}

\begin{figure}[t]
  \centering
  \includegraphics[width=\linewidth]{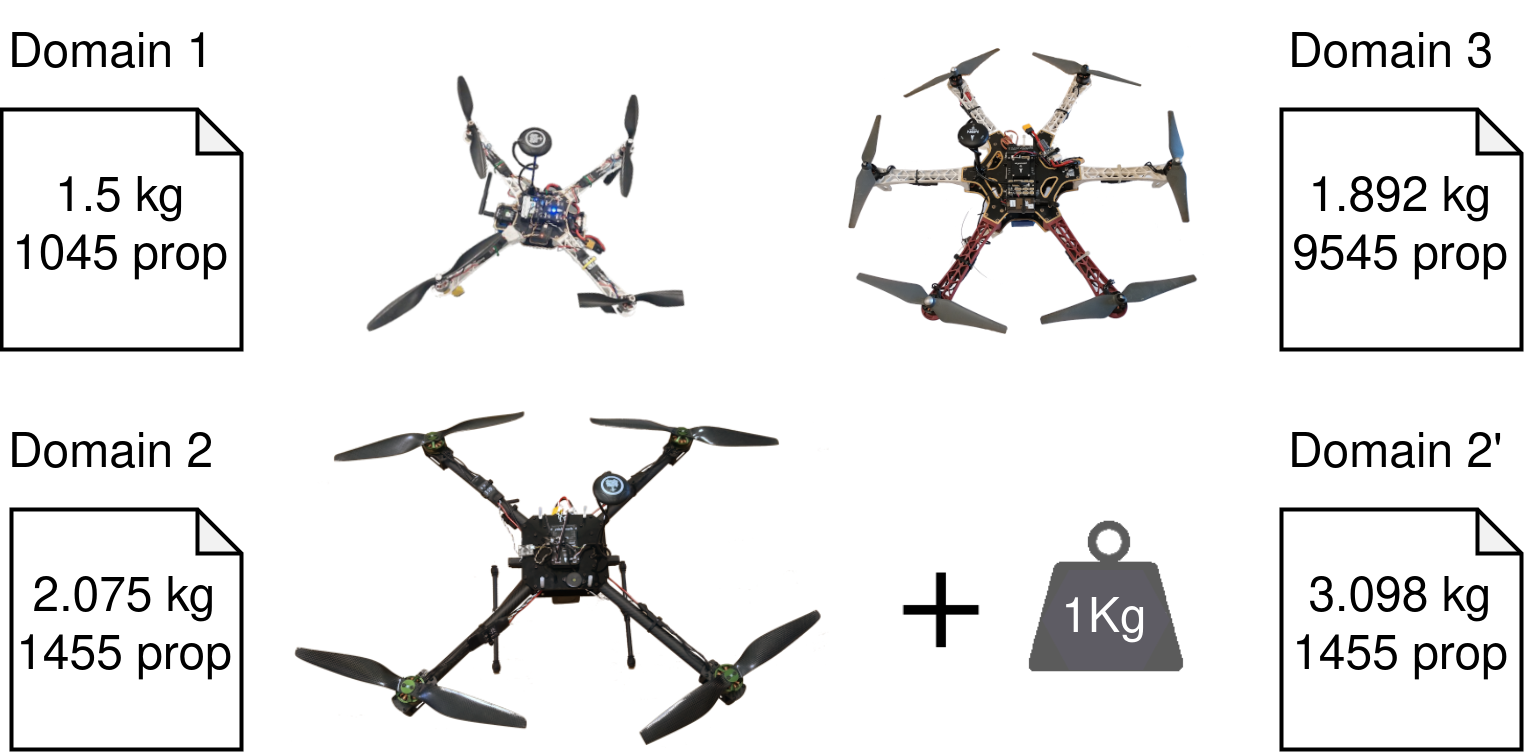}
  \caption{UAVs corresponding to each Domain.}
  \label{fig:data}
\end{figure}

To show the transfer capabilities of the proposed method, several vehicles were built to record flight data with healthy and damaged propellers. The experimental domains include different types of vehicles, and also alternate versions of the same vehicle equipped with varying payload levels, as shown in Fig.~\ref{fig:data}, and which will be described below.

\medskip
Domain~1: The first domain features a commercial quadrotor, the DJI F450, with a distance of \SI{450}{\milli\meter} between opposite motors. Each actuator consists of a DJI 2212-920KV motor paired with a 1045 plastic propeller, powered by a 4s-5000mAh LiPo battery. The total weight of the vehicle is \SI{1.5}{\kilogram}.

\medskip
Domain~2: The second domain is a custom-made quadrotor with a carbon frame. It has a distance of \SI{670}{\milli\meter} between opposite motors. The motors are Multistar 3508-268KV, coupled with 1455 carbon propellers, and powered by a 6s-5000mAh LiPo battery. The total weight of the vehicle is \SI{2.075}{\kilogram}. In order to test the robustness over different weight distributions, flight data was also gathered with the battery displaced significantly forwards and backwards.

\medskip
Domain~2': The third domain uses the same vehicle as Domain 2, but with an additional payload of \SI{1}{\kilogram}, bringing the total weight to \SI{3.098}{\kilogram}. This setup simulates a fully loaded delivery drone. An additional flight was performed with a variable weight by loading a full bottle of water with its content sloshing and leaking during the flight, as it would happen in agricultural applications.

\medskip
Domain~3: The last domain domain features a commercial hexarotor, the DJI F550, with a distance of \SI{550}{\milli\meter} between opposite motors. Each actuator set consists of a DJI 2212-920KV motor and a 9545 plastic propeller, powered by a 4s LiPo battery. The total weight of the vehicle is \SI{1.892}{\kilogram}.

\medskip

Flights were conducted with either all propellers in healthy condition or with one damaged propeller. The \textit{damage} was simulated by removing approximately 10\% of the propeller blade length from one tip. \ch{The duration of each flight in 
seconds is stated in Table~\ref{tab:classes}.}

\begin{table}[t]
\centering
\begin{tabular}{lll}
\hline
Dataset & Healthy & Damaged \\ \hline
Domain1         & 146s & 126s \\
Domain2         & 125s & 124s \\
Domain2 (Test)    & 126s & 124s \\
Domain2 (sloshing)   & 41s  & 44s  \\
Domain2 (FW)   & - & 117s \\
Domain2 (BW)   & - & 113s \\
Domain2'        & 112s & 94s  \\
Domain3         & 100s & 109s \\ \hline
\end{tabular}
\caption{Flight time for each of the vehicles, corresponding to the datasets gathered for the different experiments. }
\label{tab:classes}
\end{table}

\section{Results}

\ch{This section presents the results of the training, testing, and domain adaptations experiments performed. For clarity, a detail of which dataset is used for each experiment is presented in Table~\ref{tab:experiments}.}

\begin{table}[t]
\centering
\begin{tabular}{lll}
\hline
Experiment & Training Dataset & Testing Dataset \\ \hline
VI.A   & Domain2          & Domain2 (test)     \\
VI.B.1,2   & Domain1          & Domain2            \\
VI.B.3   & Domain1          & Domain2 (FW,BW)            \\
VI.C.1,2 & Domain2         & Domain2'           \\
VI.C.3 & Domain2          & Domain2 (sloshing)    \\
VI.D   & Domain2          & Domain3            \\ \hline
\end{tabular}
\caption{Datasets used for training and testing in each of the experiments.}
\label{tab:experiments}
\end{table}

\subsection{Classifier Accuracy on the Same Vehicle}\label{classacc}

To establish a baseline for the performance of our detection method, we first evaluated the accuracy of the neural network by training and testing on data from the same vehicle. In this experiment, inertial data was recorded from the vehicle operating in Domain~2 and processed using the previously described method to train the neural network. The model was then tested on flights conducted on a different day and under different wind conditions (Domain2 test).

All 228 samples of normal (healthy propeller) test data were correctly classified as normal, and all 228 samples of damaged propeller test data were accurately classified as damaged, resulting in 100\% classification accuracy. These results demonstrate the robustness of the detection method against environmental variations, as the data from different flights remained consistent enough to ensure perfect damage detection.

\subsection{Training and Deploying on Different Vehicles}
In this set of experiments, we train a classifier using data from Domain~1 (F450 quadrotor) and deploy it to Domain~2 (carbon fiber quadrotor). Here, inertial data is recorded from the vehicle in Domain~1 and processed to train the neural network.

After training the network, the test dataset of Domain~2 is used for evaluating the performance, including as before 228 samples of healthy propeller data and 228 samples of damaged propeller data.

\subsubsection{Naive Approach}\label{naive1}
\begin{figure}[t]%
    \centering
    \begin{subfigure}{0.48\linewidth}
        \includegraphics[width=\linewidth]{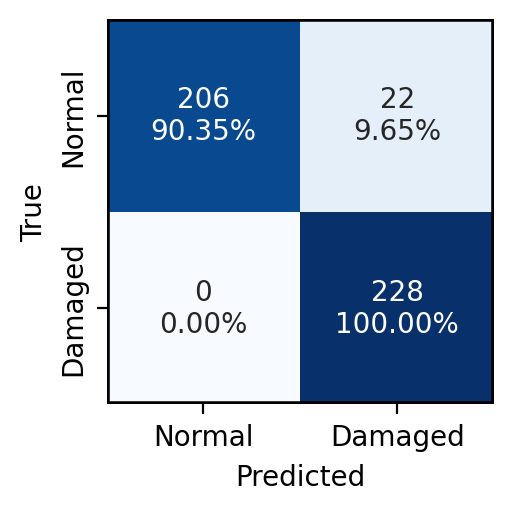}
        \caption{Naive approach}\label{fig:cm_naive}
    \end{subfigure}
    \hfill
    \begin{subfigure}{0.48\linewidth}
        \includegraphics[width=\linewidth]{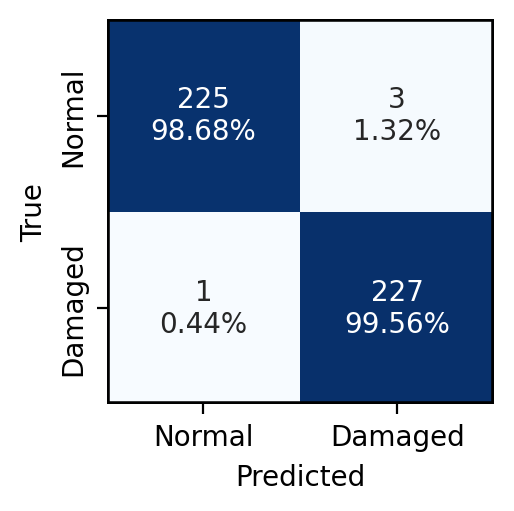}
        \caption{Proposed adaptation method}\label{fig:cm_g2c}%
    \end{subfigure}
    \caption{Confusion matrix: trained in Domain~1 and tested in Domain~2. Accuracy increases from 0.9518 to 0.9912 using the domain adaptation method.}%
\end{figure}

In the naive approach, the trained network is directly deployed to Domain~2 without any additional adjustments. As shown in Fig.~\ref{fig:cm_naive}, this approach does not perform well, particularly in terms of precision. In a real-world scenario, this could lead to unnecessary flight interruptions or premature maintenance, both of which could be costly.

\subsubsection{Domain Adaptation} \label{sec:dacarbono}
In the domain adaptation approach, the network trained on Domain~1 data is deployed to Domain~2, but the data collected from Domain~2 is processed further, as described in Section~\ref{domadapt}. After obtaining the Welch power spectrum for each one-second sample, a domain adaptation factor \(s_{2 \rightarrow 1} = 0.69\) is applied.

The results, shown in Fig.~\ref{fig:cm_g2c}, demonstrate a significant improvement in both precision and accuracy compared to the naive approach.

\subsubsection{Change in Weight Distribution}

To assess the method's sensitivity to changes in weight distribution, the Domain 2 data with the battery shifted (ForWard - FW  and BackWards - BW) was used. The same experiment described in Section~\ref{sec:dacarbono} was then repeated. This shift altered the propeller speeds and affected the vibration patterns, leading to changes in the power spectrum. The network, trained on balanced data from Domain~1, exhibited reduced detection performance when the weight was shifted, particularly when the weight was moved away from the compromised propeller. The accuracy results are shown in Table~\ref{tab:dist}.

\begin{table}[t]
\centering
\begin{tabular}{llll}
\hline
Shift & Accuracy &  Recall & Precision \\ \hline
No shift & \textbf{0.9912} & \textbf{0.9956} & \textbf{0.9870} \\
Forward    & 0.9680 & 0.9949 & 0.9680 \\
Backwards     & 0.9330 & 0.9703 & 0.8909 \\
\hline
\end{tabular}
\caption{Performance statistics of Domain~1 to Domain~2 adaptation with weight distribution shifts.}
\label{tab:dist}
\end{table}

\subsection{Training and Deploying on the Same Vehicle with Different Weight}

In this set of experiments, we train the damage detector using flight data from Domain~2 (the carbon fiber quadrotor) and test it using data from Domain~2', which is the same vehicle but with an additional \SI{1}{\kilogram} of weight.

\subsubsection{Naive Approach}

\begin{figure}[t]%
    \centering
    \begin{subfigure}{0.48\linewidth}
        \includegraphics[width=\linewidth]{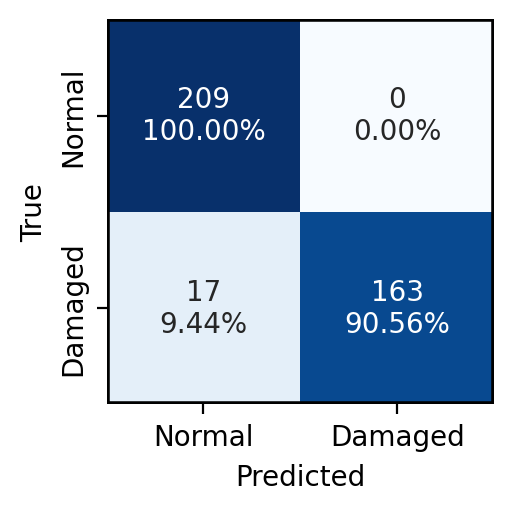}
        \caption{Naive approach}\label{fig:naive_cm_c2c}
    \end{subfigure}
    \hfill
    \begin{subfigure}{0.48\linewidth}
        \includegraphics[width=\linewidth]{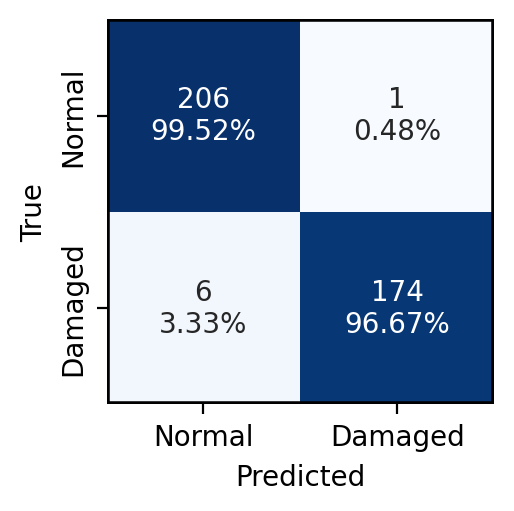}
        \caption{Proposed adaptation method}\label{fig:da_cm_c2c}%
    \end{subfigure}
    \caption{Confusion matrix: trained on Domain~2 and tested on Domain~2'. Accuracy improves from 0.9528 to 0.9809 with the domain adaptation method.}%
\end{figure}

As in Section~\ref{naive1}, the simplest approach is to train the model on the source dataset and deploy it to the target dataset without any adjustments. As shown in the confusion matrix in Fig.~\ref{fig:naive_cm_c2c}, this method results in lower recall, meaning potential failure to detect damage during flight or missing necessary maintenance.

\begin{table}[t]
\centering
\begin{tabular}{lcc}
\hline
\multirow{2}{*}{Extra Weight} & \multicolumn{2}{c}{Accuracy} \\ 
                              & Naive         & Adapted        \\ \hline
+0Kg                          & 1.0           & -              \\
+0.8$\sim$0Kg (sloshing)      & 0.9957        & \textbf{0.9975}         \\
+1Kg                          & 0.9528        & \textbf{0.9809}         \\ \hline
\end{tabular}
\caption{Performance statistics of Domain~2 adaptation with liquid and solid extra weight.}
\label{tab:dist2}
\end{table}

\subsubsection{Domain Adaptation}
Similar to the previous experiments, domain adaptation is applied to account for the weight difference between the two vehicles, which affects the nominal motor frequency. As seen in Fig.~\ref{fig:da_cm_c2c}, recall improves from 0.9056 in the naive approach to 0.9667, with the number of misclassified damaged samples reduced by two-thirds.

\subsubsection{Time-Variable Weight}
One application of this system is for UAVs with changing weight, such as agricultural or delivery drones. In this experiment, the vehicle from Domain~2 carries a water reservoir that empties over time, starting with an additional \SI{0.8}{\kilogram} (40\% extra weight) and ending with none.  Using this method, accuracy improves during the first third of the flight, when the load is heaviest, as shown in Table~\ref{tab:dist2}, and reaches perfect detection by the end, similar to the results in Section~\ref{classacc}.

\subsection{Different Vehicle Configurations}
The final test evaluated how well the proposed method generalizes when the two domains are quite different. The network trained on Domain~1 was deployed on Domain~3, using both the naive and domain adaptation approaches. Given that the vehicles were from the same manufacturer, with similar motors and the same flight controller, initial expectations were that the method might perform well. However, the results were poor, with the network consistently classifying 100\% of the flight as either healthy or damaged, depending on whether domain adaptation was applied, regardless of the actual state of the propellers.

\begin{figure}[t]
  \centering
  \includegraphics[width=\linewidth]{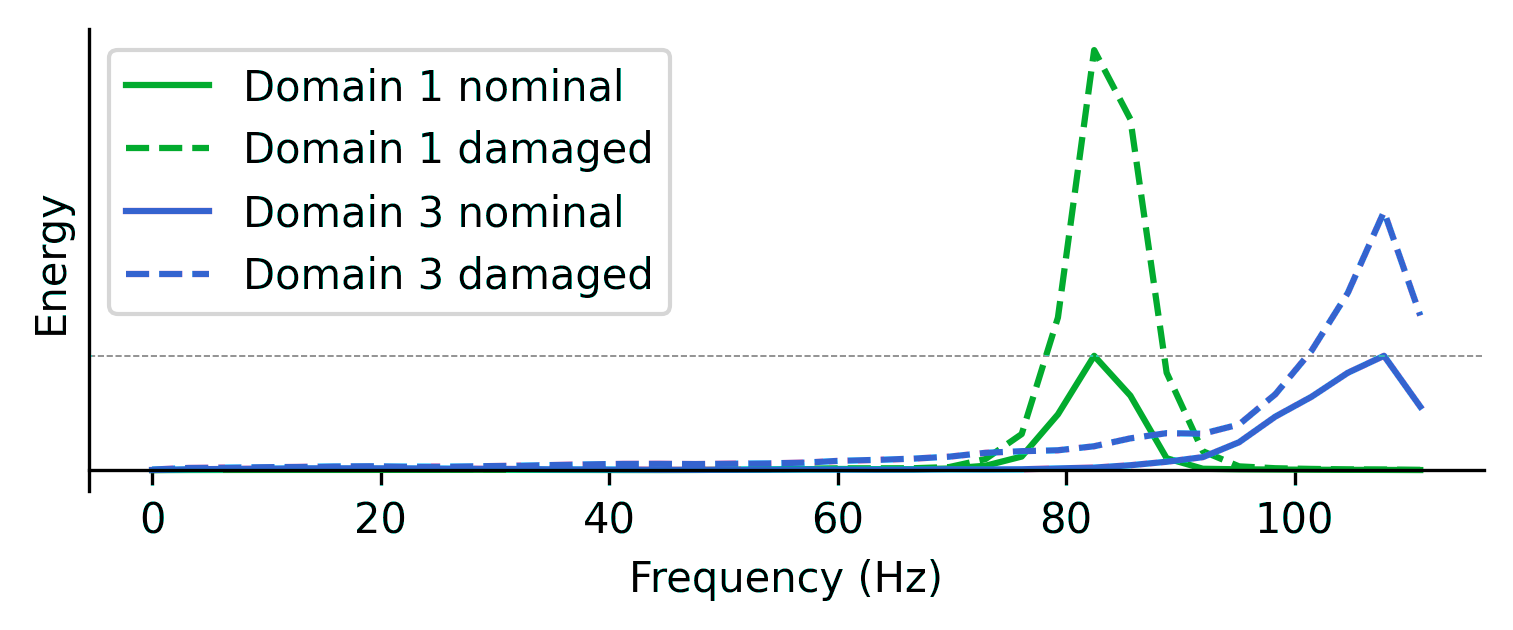}
  \caption{Normalized power spectra for Domains 1 and 3 show that the quadrotor (Domain 1) is more sensitive to equal single propeller damage than the hexarotor (Domain 3).}
  \label{fig:dam_sens}
\end{figure}

Figure~\ref{fig:dam_sens} shows the normalized power spectrum from one of the accelerometers for both domains, aligned to the energy peak of healthy flights. In Domain~3, the energy peak in the damaged case is smaller than in Domain~1, indicating that the hexarotor is less affected by propeller damage. This may be due to several factors: the hexarotor's structure could absorb vibrations differently, and single propeller damage affects only \(\frac{1}{6}\) of the motors in a hexarotor, compared to \(\frac{1}{4}\) in a quadrotor, which could result in higher relative vibrations.
\ch{The reason why the method fails, or more generally why the spectra changes in this configuration, is an interesting avenue for future research.}

\subsection{Computational Efficiency} In \cite{pose2023}, real-time tests achieved 42.29 inferences per second on 1-second data windows. By enhancing the algorithm and eliminating the need for a filter bank, our method now achieves 310.83 inferences per second on the same window size, a 7.35x speedup. The time bottleneck has shifted from feature generation, which we optimized, to neural network inference, which can be easily accelerated with an embedded GPU. These results, tested on a Raspberry Pi 4B (4GB) running Python 3.12.6 and Torch 2.4.1, show that this system can run online on a UAV.

\section{Discussion}
\label{dis}

The experimental results show that thew proposed method achieves a good performance in detecting damaged propellers, and is robust against structure and motor-propeller variations, while only requiring basic information such as vehicle's weight and actuators' performance. The methodology improves precision over naive approaches, and the results are better when applied between vehicles with the same base structure, such as quadrotors. 

However, it presents limitations when applied between structures that differ in actuator number and/or distribution, such as the hexarotor case. Still, only this particular hexarotor has been tested here, and further and more detailed studies will be made in the future to address a wider variety of multirotors, considering different number of rotors and motors' distributions.

When considering similar structures, it is remarkable that the method is able to reach high accuracy even when the characteristics are modified during flight, as shown for the time-variable weight case. 

\section{Conclusions and future work}
\label{con}

This study presented a method grounded in first principles to adapt propeller damage detection models for different UAVs. We showed how physical changes, particularly weight, affect the power spectra from a vehicle's inertial sensors, offering insights into how various UAVs respond to propeller damage.
Quadrotors and hexarotors respond differently to damage, emphasizing the need for UAV-specific detection strategies. We also improved an existing fault detection algorithm, enhanced computational efficiency.

Future research will focus on more robust methods that can handle larger domain shifts, such as adapting between quadrotors and hexarotors, where the current method has shown limitations. Further exploration will be made in the incidence of damaged propellers when there are different number of motors, and/or when the distribution is not the same.

Also, in line with some of the authors' previous work, the proposed solution will be extended not only to consider detection of different types and magnitudes of damage, but also to be able to identify the type of damage and the location of the motor affected. To this end, a dual approach including regression networks, to quantify the damage, in addition to classifiers will be implemented.

\section*{Acknowledgments}

This work has been partially supported by PICT-2019-0373 project from Agencia Nacional de Investigaciones Cient\'ificas y Tecnol\'ogicas, and UBACyT-0421BA project from the University of Buenos Aires (UBA).

\bibliographystyle{IEEEtran}
\bibliography{biblio}
\balance

\end{document}